\documentclass{article} 
\usepackage{iclr2027_conference,times}

\usepackage{hyperref}
\usepackage{url}

\usepackage[T1]{fontenc}
\usepackage[utf8]{inputenc}
\usepackage{microtype}
\usepackage{graphicx}
\usepackage{amsmath}
\usepackage{amssymb}
\usepackage{float}
\usepackage{threeparttable}
\usepackage{booktabs}
\usepackage{multirow}

\title{Learning What to Distill: Bilevel Top-$K$ Token Selection for Self-Distillation in Large Language Models}

\author{
Heng Liang \\
Department of Computer and Information Sciences \\
Temple University \\
\texttt{heng.liang@temple.edu}
\And
Xinwen Zhang \\
Department of Computer and Information Sciences \\
Temple University \\
\texttt{ellenz@temple.edu}
\And
Hongchang Gao \\
Department of Computer and Information Sciences \\
Temple University \\
\texttt{hongchang.gao@temple.edu}
}

\begin{document}

\maketitle


\begin{abstract}
Large language models have shown strong reasoning capabilities, but their high inference costs make knowledge distillation an important approach for transferring such capabilities to compact models in resource-constrained scenarios.
On-policy self-distillation further reduces the reliance on external large teacher models while improving the reasoning ability of compact language models.
However, existing methods typically either distill all token positions uniformly or select tokens using fixed heuristic criteria, assigning the same distillation strength to the selected positions rather than adaptively learning which tokens are most beneficial for distillation. To address these limitations, we propose \textbf{BiToK-SD}
(\textbf{Bi}level \textbf{To}p-$\mathbf{K}$ Token Selection for
\textbf{S}elf-\textbf{D}istillation), a bilevel-optimization-based token selection method that learns where distillation should be applied during on-policy self-distillation.
Specifically, BiToK-SD is formulated as a bilevel optimization problem, where the lower-level problem models Top-$K$ token selection as a differentiable threshold-based relaxation, allowing the selected positions to adapt as the student policy evolves, while the upper-level problem performs knowledge distillation on the selected positions.
Experiments on mathematical reasoning benchmarks show that BiToK-SD achieves
the best average performance among all compared methods while requiring only
lightweight additional computation.
\end{abstract}

\section{Introduction}

Large language models (LLMs) have achieved remarkable progress on complex reasoning tasks~\citep{wei2022chain}.
However, their large model sizes and high inference costs make them difficult to deploy in resource-constrained scenarios.
Knowledge distillation (KD) provides an effective way to transfer the capabilities of stronger models to smaller ones, and has therefore been widely used to improve the reasoning ability of compact LLMs~\citep{xu2024survey}.
Existing LLM distillation methods can be broadly categorized into off-policy and on-policy approaches: off-policy methods rely on fixed teacher outputs or pre-collected reasoning trajectories~\citep{hsieh2023distilling,shridhar2023distilling}, while on-policy methods construct training signals from the student's current rollouts to better match the student distribution~\citep{agarwal2024policy}.
Despite their effectiveness, these methods often require either high-quality offline data, a strong external teacher model, or an online reward/teacher model, making the overall training process costly and less flexible.

To reduce the dependence on external large teacher models, recent studies have begun to explore on-policy self-distillation~\citep{shenfeld2026self,hubotter2026reinforcement}.
Representative methods such as OPSD construct feedback-conditioned token-level correction signals from the model's own generated reasoning trajectories, enabling the model to distill from itself~\citep{zhao2026self}.
Such methods remove the need for a large teacher model and provide an efficient way to improve smaller reasoning models.
However, they typically apply the same distillation objective uniformly across all token positions in the generated trajectory.
This uniform treatment overlooks the heterogeneous roles of tokens in long chain-of-thought reasoning, where tokens may correspond to exploration, correction, reflection, or final answer derivation.

A natural improvement is selective token-level distillation, where KD signals are applied only to token positions that are more important or more suitable for distillation.
Recent studies have explored token selection or reweighting based on predefined indicators, such as token entropy, teacher--student divergence, reverse KL divergence, or their combinations such as Soft-OR scores~\citep{tavor2026rethinking,xu2026tip}.
However, because these methods rely on predefined token-level scores, their selection criteria remain fixed throughout training and are not adapted to the self-distillation objective. Since the student distribution changes continuously during on-policy self-distillation, fixed token-selection scores may fail to consistently identify token positions that are beneficial for distillation.
Thus, the key question is not only whether token selection is needed, but how to learn adaptive token-level weights that determine both the selected positions and their distillation strengths.

Learning token selection is challenging because Top-$K$ selection is inherently discrete and non-differentiable.
Prior work on X-risk minimization and learning to rank has shown that Top-$K$-related objectives often require specialized optimization techniques, including bilevel formulations and differentiable relaxations of ranking or sorting operations~\citep{yang2022algorithmic,swezey2021pirank}.
Inspired by this perspective, we formulate token selection as a learnable differentiable Top-$K$ problem, enabling the selection mechanism to be optimized with respect to the self-distillation objective.

To this end, we propose BiToK-SD, a bilevel-optimization-based Top-$K$ token selection method for on-policy self-distillation.
Unlike uniform distillation or heuristic token selection based on fixed importance scores, BiToK-SD learns adaptive token-level distillation weights through bilevel optimization, allowing the model to determine both where and how strongly distillation should be applied.
We evaluate BiToK-SD on mathematical reasoning benchmarks under thinking-mode evaluation.
Experimental results show that BiToK-SD achieves the best average
performance among all compared methods while requiring only lightweight
additional computation.
Further analyses and ablation studies show that BiToK-SD preserves stronger policy exploration, induces more reflection-style reasoning behavior, and remains robust across different Top-$K$ ratios.

\section{Related Work}

Knowledge distillation (KD), following the classical teacher--student paradigm~\citep{hinton2015distilling} and its sequence-level extensions~\citep{kim2016sequence}, has been widely used to transfer the reasoning and instruction-following abilities of large language models (LLMs) to smaller models~\citep{gu2024minillm, agarwal2024policy, ko2024distillm, shing2025taid, kim2024promptkd, guo2025deepseek}.
Existing KD methods can be categorized into off-policy and on-policy approaches.
Off-policy KD relies on fixed teacher outputs, offline teacher-generated responses, or pre-collected reasoning trajectories~\citep{hsieh2023distilling, shridhar2023distilling, li2023symbolic, chen2023disco, jiang2023lion}, which may cause distribution mismatch when the student is trained on static trajectories rather than its own rollouts.
On-policy KD alleviates this issue by constructing supervision from student-generated trajectories, better matching the evolving student distribution~\citep{shao2024deepseekmath, ye2025black, ye2026policy, ko2026scaling}.

More recently, self-distillation (SD) further reduces reliance on external teacher models by allowing the model to learn from its own generated reasoning trajectories~\citep{chen2024self, hubotter2026reinforcement, sang2026policy, shenfeld2026self}.
For example, \citet{zhao2026self} propose OPSD, where a single model serves as both teacher and student during on-policy self-distillation.
Since uniformly distilling all generated tokens may introduce redundant or even harmful supervision, recent work has begun to explore selective token-level distillation methods~\citep{jung2025todi, kim2026explain, huang2025selectkd}.
\citet{xu2026tip} propose Soft-OR scores to select informative tokens for distillation based on student uncertainty and teacher--student divergence.

However, most existing selective distillation methods formulate token selection as a binary decision, assigning the same distillation weight to all selected positions without explicitly modeling their varying importance.
This limitation suggests the need for more flexible token-level weighting mechanisms beyond binary selection.

A related line of work studies differentiable optimization for ranking and top-$K$ selection.
Many ranking-oriented objectives, such as NDCG, MAP, and precision at top-$K$, depend on discrete ranks or top-$K$ selection and are therefore difficult to optimize directly with gradient-based methods~\citep{blondel2020fast, yang2022algorithmic}.
To address this issue, prior work has developed differentiable relaxations of sorting and ranking operators, such as PiRank~\citep{swezey2021pirank} and NeuralNDCG~\citep{pobrotyn2021neuralndcg}, as well as bilevel or compositional formulations for optimizing top-$K$ ranking metrics such as NDCG~\citep{qiu2022large}.
While these methods provide useful tools for differentiable approximations of discrete top-$K$ selection, they are designed for conventional ranking, retrieval, or recommendation tasks rather than adaptive token-level weighting in LLM self-distillation.

\section{Method}
\label{sec:method}

\subsection{On-Policy Self-Distillation Setup}

We follow OPSD~\citep{zhao2026self} to construct token-level self-distillation signals from the model's own on-policy reasoning trajectories.
Given an input problem $x_i$, the current student policy generates a reasoning trajectory
$\hat{y}_i=(\hat{y}_1^{(i)},\ldots,\hat{y}_{L_i}^{(i)})$,
where $L_i$ denotes the length of the generated trajectory.
Since distillation is performed on next-token distributions, the trajectory provides $T_i=L_i-1$ distillation positions, indexed by $t\in\{1,\ldots,T_i\}$.
For each position $t$, we denote the student distribution and the feedback-conditioned self-teacher distribution over the next token as
$p_{S,t}^{(i)}$ and $p_{T,t}^{(i)}$, respectively.
Here, the self-teacher distribution is conditioned on the reference or feedback-providing solution used by the OPSD-style teacher context.

We formulate the token-level self-distillation signal as a full-vocabulary distribution divergence between the feedback-conditioned self-teacher and the student.
In our experiments, we instantiate this divergence as the teacher-to-student KL divergence.
For each token position $t$ in the $i$-th training example, the per-position loss is defined as
\begin{equation}
\ell_{\mathrm{pos}}^{(i)}(t;\theta)
=
\mathrm{KL}
\left(
p_{T,t}^{(i)} \middle\| p_{S,t}^{(i)}
\right).
\label{eq:pos-loss}
\end{equation}
This full-vocabulary objective provides dense token-level feedback by reducing the divergence between the student distribution and the feedback-conditioned self-teacher distribution over the entire vocabulary.

\subsection{Differentiable Top-$K$ Token Weighting}

Recent token-selection methods commonly compute a token-level importance score and then apply a hard Top-$K$ selector to retain only the highest-scoring positions.
However, this strategy is limited by its discrete nature and its uniform treatment of selected tokens.
Specifically, the binary Top-$K$ operation is non-differentiable, and all selected tokens receive the same weight, ignoring that different tokens may contribute to distillation with different strengths.
BiToK-SD addresses these limitations by replacing hard Top-$K$ selection with a differentiable threshold-based relaxation, which yields adaptive soft token weights instead of a binary mask.

For each valid token position $(i,t)$ in a mini-batch $\mathcal{B}$, we compute a token-level utility score $u_t^{(i)}$ to estimate how suitable this position is for distillation.
In our main experiments, we use the Soft-OR score as the default token-level importance score:
\begin{equation}
u_t^{(i)}
=
\hat{h}_t^{(i)}
+
\hat{d}_t^{(i)}
-
\hat{h}_t^{(i)}\hat{d}_t^{(i)},
\label{eq:softor-utility}
\end{equation}
where $h_t^{(i)}=H(p_{S,t}^{(i)})$ is the student entropy,
$d_t^{(i)}=\mathrm{KL}(p_{S,t}^{(i)}\|p_{T,t}^{(i)})$ is the reverse KL divergence, and hats denote min--max normalization over valid token positions in the mini-batch.
The score serves as a token-level importance proxy and is recomputed at each training step as the student policy evolves.
BiToK-SD is not tied to this specific score and can also use other token-level indicators, such as student entropy, teacher entropy, forward KL divergence, or reverse KL divergence.

Let $\Omega_{\mathcal{B}}$ denote the set of valid non-padding token positions in mini-batch $\mathcal{B}$, and let
$M_{\mathcal{B}}=|\Omega_{\mathcal{B}}|$.
Given a target Top-$K$ ratio $k$, we set
\begin{equation}
K_{\mathcal{B}}
=
\max\left(1,\left\lceil kM_{\mathcal{B}}\right\rceil\right),
\end{equation}
where $K_{\mathcal{B}}$ denotes the target number of high-utility positions in the mini-batch.

A hard batch-level Top-$K$ selector can be written as
\begin{equation}
z_t^{(i)}
=
\mathbf{1}\!\left(
u_t^{(i)}
\in
\mathrm{TopK}
\left(
\{u_{t'}^{(i')}\}_{(i',t')\in\Omega_{\mathcal{B}}}
\right)
\right).
\label{eq:hard-topk}
\end{equation}
However, this binary selector is non-differentiable and assigns the same weight to all selected positions.
As a result, it cannot express different distillation strengths among selected tokens.
BiToK-SD therefore replaces the hard selector with a differentiable threshold-based relaxation.

The relaxed threshold is obtained as
\[
\hat{\rho}_{\mathcal{B}}
\approx
\operatorname*{arg\,min}_{\rho\in\mathbb{R}}
\Psi_{\mathcal{B}}(\rho;u),
\]
where
\begin{equation}
\Psi_{\mathcal{B}}(\rho;u)
= \frac{K_{\mathcal{B}}+\varepsilon}{M_{\mathcal{B}}}\rho
+ \frac{\tau_2}{2}\rho^2
+ \frac{\tau_1}{M_{\mathcal{B}}}
\sum_{(i,t)\in\Omega_{\mathcal{B}}}
\log\!\left(1+\exp\!\left(\frac{u_t^{(i)}-\rho}{\tau_1}\right)\right).
\label{eq:smooth-threshold}
\end{equation}
Here, $\tau_1$ controls the smoothness of the softplus approximation, $\tau_2$ stabilizes the threshold optimization, and $\varepsilon$ avoids boundary ambiguity.
In practice, we approximate $\hat{\rho}_{\mathcal{B}}$ using a warm-started momentum gradient update and detach the resulting threshold before computing the soft mask.

Given $\hat{\rho}_{\mathcal{B}}$, the soft token-selection mask is defined as
\begin{equation}
\tilde{z}_t^{(i)}
=
\sigma\!\left(
\frac{u_t^{(i)}-\hat{\rho}_{\mathcal{B}}}{\tau_m}
\right)
\cdot
\mathbf{1}\!\left((i,t)\in\Omega_{\mathcal{B}}\right),
\label{eq:soft-mask}
\end{equation}
where $\tau_m$ controls the sharpness of the mask.
Unlike hard Top-$K$ selection, $\tilde{z}_t^{(i)}$ is continuous and assigns different strengths to different token positions.
Thus, BiToK-SD determines not only which tokens are emphasized, but also how strongly they contribute to the distillation objective.

\subsection{Weighted Self-Distillation via Bilevel Optimization}

The relaxed threshold in Eq.~\eqref{eq:smooth-threshold} is obtained from an inner optimization problem, while the student model is updated by minimizing a weighted self-distillation objective.
This gives the following approximate bilevel formulation:
\begin{equation}
\min_{\theta}\ \mathcal{L}_{\mathrm{BiToK\text{-}SD}}(\theta;\hat{\rho}_{\mathcal{B}})
\quad \mathrm{s.t.} \quad
\hat{\rho}_{\mathcal{B}}
\approx
\operatorname*{arg\,min}_{\rho}
\Psi_{\mathcal{B}}(\rho;u).
\label{eq:bilevel}
\end{equation}
Here, $\Psi_{\mathcal{B}}$ denotes the relaxed threshold objective in Eq.~\eqref{eq:smooth-threshold}.
The approximation reflects that the threshold is updated by a warm-started gradient step rather than solved exactly at every iteration.

Given the soft mask $\tilde{z}_t^{(i)}$, BiToK-SD minimizes the batch-normalized weighted self-distillation loss:
\begin{equation}
\mathcal{L}_{\mathrm{BiToK\text{-}SD}}(\theta)
=
\frac{
\sum_{(i,t)\in\Omega_{\mathcal{B}}}
\tilde{z}_t^{(i)}
\ell_{\mathrm{pos}}^{(i)}(t;\theta)
}{
\sum_{(i,t)\in\Omega_{\mathcal{B}}}
\tilde{z}_t^{(i)}
+\delta
},
\label{eq:bitok-loss}
\end{equation}
where $\delta$ is a small constant for numerical stability.

Equivalently, we define the normalized token-level distillation weight as
\begin{equation}
\bar{w}_t^{(i)}
=
\frac{
\tilde{z}_t^{(i)}
}{
\sum_{(i',t')\in\Omega_{\mathcal{B}}}
\tilde{z}_{t'}^{(i')}
+\delta
},
\label{eq:normalized-weight}
\end{equation}
and rewrite Eq.~\eqref{eq:bitok-loss} as
\begin{equation}
\mathcal{L}_{\mathrm{BiToK\text{-}SD}}(\theta)
=
\sum_{(i,t)\in\Omega_{\mathcal{B}}}
\bar{w}_t^{(i)}
\ell_{\mathrm{pos}}^{(i)}(t;\theta).
\label{eq:weighted-loss}
\end{equation}

This formulation does not modify the underlying token-level self-distillation loss, but instead changes how the loss is weighted across token positions.
For stability and efficiency, the token weights are treated as stop-gradient coefficients within each update.
Nevertheless, because the utilities and threshold are recomputed from the evolving student policy during on-policy training, the emphasized token positions and their effective weights can change dynamically across training steps.

\section{Experiments}

\subsection{Experimental Setup}

\textbf{Models and Data.}
We conduct on-policy self-distillation experiments at three model scales,
using Qwen3-1.7B, Qwen3-4B, and Qwen3-8B as base models.
At each scale, all methods are initialized from the same checkpoint and
trained on the same data to ensure a fair comparison.
We train on the mathematical reasoning subset of
OpenThoughts~\citep{guha2025openthoughts}, which contains approximately 30K
problem--solution pairs with chain-of-thought reasoning trajectories.

\textbf{Baselines.}
We compare BiToK-SD with some representative baselines covering supervised fine-tuning, reinforcement learning, on-policy self-distillation, and heuristic token selection.
SFT performs standard supervised fine-tuning on expert reasoning trajectories.
GRPO~\citep{shao2024deepseekmath} optimizes the policy using binary outcome rewards and serves as a reinforcement-learning baseline.
OPSD~\citep{zhao2026self} performs on-policy self-distillation by constructing feedback-conditioned token-level correction signals from the model's own rollouts.
TIP~\citep{xu2026tip} selects the top-$50\%$ positions by the same Soft-OR score
(Eq.~\eqref{eq:softor-utility}) with a hard mask and, following the original
method, optimizes the reverse KL divergence $\mathrm{KL}(p_S\|p_T)$ on the
selected positions.

\textbf{Evaluation and Implementation.}
We evaluate on AIME24, AIME25, and HMMT25 under thinking mode across three
model scales (Qwen3-1.7B, 4B, and 8B). 
Since AIME24, AIME25, and HMMT25 each contain only a small number of
problems, single-sample accuracy is subject to high variance. We therefore
report Avg@64, the mean accuracy over 64 independently sampled solutions per
problem.
All methods are trained for 200 steps and evaluated at the final checkpoint;
we deliberately avoid test-set-based checkpoint selection.

\textbf{Hyperparameters.}
We fine-tune all methods with AdamW using a learning rate of
$5\!\times\!10^{-6}$, a per-device batch size of $2$ with gradient
accumulation of $2$ over $4$ GPUs (effective batch size $16$), a maximum
completion length of $1024$ tokens, and a sampling temperature of $1.1$.
For the 8B model, we use a per-device batch size of $1$ with gradient
accumulation of $4$ to fit 48GB GPUs, keeping the effective batch size at $16$.
Parameter-efficient fine-tuning is performed with LoRA
($r=64$, $\alpha=128$) applied to all attention and MLP projections.
For BiToK-SD, we adopt the Soft-OR score as the default token-level importance
score and set the top-$K$ ratio to $k=50\%$ of generated tokens.

For OPSD and BiToK-SD, we use the full-vocabulary teacher-to-student
KL divergence as the token-level distillation objective.
For the differentiable threshold relaxation, we set the softplus temperature
to $\tau_1=0.1$, the quadratic regularization coefficient to $\tau_2=0.01$,
and the mask temperature to $\tau_m=0.1$.
We set $\varepsilon=10^{-6}$ to avoid boundary ambiguity in
Eq.~\eqref{eq:smooth-threshold} and $\delta=10^{-6}$ for numerical stability
in Eq.~\eqref{eq:bitok-loss}.
The smooth threshold $\rho$ is updated by an inner threshold solver with
step size $\eta=0.1$, momentum coefficient $\gamma=0.9$, and
variance-reduction coefficient $\beta_{\rho}=0.1$.

Unless otherwise stated, all baselines share the same optimizer, learning rate,
effective batch size, number of training steps, LoRA configuration, and evaluation
decoding settings as BiToK-SD. For GRPO, we use a group size of $G=8$ rollouts per
prompt (two prompts per rollout batch), a maximum completion length of 8{,}192 tokens, a
sampling temperature of $1.2$, two optimization iterations per rollout batch, and
no KL penalty, so that GRPO processes 200 distinct prompts over 200 steps. For SFT,
we use a maximum sequence length of 20{,}000 tokens. 

\subsection{Main Results}

Table~\ref{tab:main-results} reports Avg@64 accuracy under thinking mode at the
final checkpoint (200 steps) across three model scales. Overall, BiToK-SD
achieves the highest average performance at all three scales, demonstrating the
effectiveness of bilevel token selection for on-policy self-distillation.
Compared with the base model, SFT degrades performance on every benchmark and
at every scale despite using ground-truth solutions, suggesting that directly
imitating fixed trajectories is less effective in our on-policy reasoning
setting. GRPO improves over the base model at 1.7B (+1.10 average points), but
its advantage disappears at larger scales, performing on par with the base
model at 4B and slightly below it at 8B. This suggests that outcome-level
rewards provide only coarse supervision while requiring substantially higher
training cost due to group sampling.

OPSD is the strongest baseline and uses feedback-conditioned correction
signals for on-policy self-distillation, improving over the base model at all
three scales. However, it still distills over all token positions.
TIP, which uses the same Soft-OR score with a hard Top-$K$ mask and the reverse KL
objective of its original formulation, falls below OPSD in all nine
scale--benchmark combinations. Since TIP differs from BiToK-SD in both the
selection rule and the divergence, this gap cannot be attributed to either factor
alone; \citet{zhao2026self} likewise report that reverse KL underperforms forward
KL in on-policy self-distillation.
In contrast, BiToK-SD constructs adaptive soft token weights through a bilevel
threshold relaxation, enabling more flexible token-level distillation. As a
result, it further improves over OPSD by 0.63, 0.58, and 0.21 average points at
1.7B, 4B, and 8B, respectively, and obtains the best result in seven of the nine
scale--benchmark cells.

\begin{table}[t]
\centering
\small
\caption{Main results across three model scales. We report Avg@64
accuracy (\%) under thinking mode, evaluated at the final checkpoint
(200 steps) for all methods. Best result per scale and benchmark in
bold; ties are bolded jointly.}
\label{tab:main-results}
\setlength{\tabcolsep}{15pt}
\renewcommand{\arraystretch}{1.10}
\begin{center}
\begin{tabular}{llcccc}
\toprule
Scale & Method & AIME24 & AIME25 & HMMT25 & Avg. \\
\midrule
\multirow{6}{*}{1.7B}
 & Base     & 48.07 & 35.62 & 23.23 & 35.64 \\
 & SFT      & 33.44 & 25.73 & 15.68 & 24.95 \\
 & GRPO     & 49.74 & 36.98 & 23.49 & 36.74 \\
 & OPSD     & 55.10 & 39.69 & 27.76 & 40.85 \\
 & TIP      & 48.12 & 35.52 & 23.49 & 35.71 \\
 & \textbf{BiToK-SD} & \textbf{55.99} & \textbf{40.21} & \textbf{28.23} & \textbf{41.48} \\
\midrule
\multirow{6}{*}{4B}
 & Base     & 73.39 & 66.09 & 43.12 & 60.87 \\
 & SFT      & 59.90 & 49.43 & 30.73 & 46.69 \\
 & GRPO     & \textbf{74.53} & 65.10 & 43.80 & 61.14 \\
 & OPSD     & 73.54 & 67.40 & 43.54 & 61.49 \\
 & TIP      & 73.07 & 63.12 & 42.19 & 59.46 \\
 & \textbf{BiToK-SD} & 73.96 & \textbf{67.71} & \textbf{44.53} & \textbf{62.07} \\
\midrule
\multirow{6}{*}{8B}
 & Base     & 76.72 & 68.28 & 45.26 & 63.42 \\
 & SFT      & 74.06 & 65.57 & 43.23 & 60.95 \\
 & GRPO     & 76.15 & 68.18 & 44.64 & 62.99 \\
 & OPSD     & 76.67 & 70.26 & \textbf{46.98} & 64.64 \\
 & TIP      & 75.57 & 66.67 & 44.22 & 62.15 \\
 & \textbf{BiToK-SD} & \textbf{77.14} & \textbf{70.47} & 46.93 & \textbf{64.85} \\
\bottomrule
\end{tabular}
\end{center}
\end{table}

\begin{figure}[t]
\begin{center}
\includegraphics[width=\linewidth]{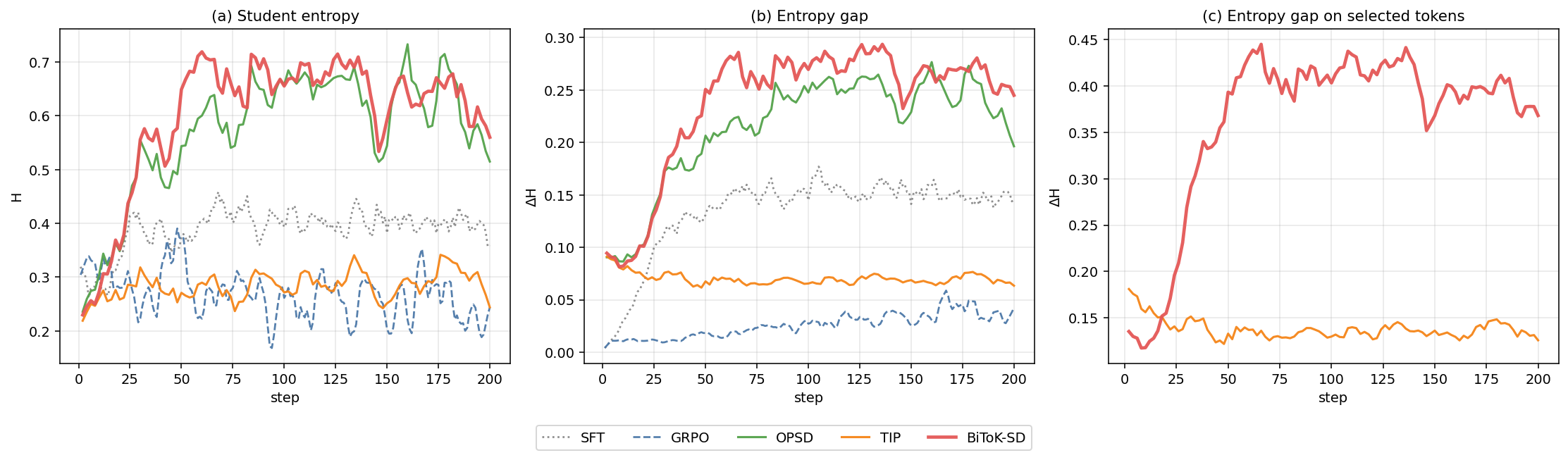}
\end{center}
\caption{
Exploration dynamics during training.
(a) Student entropy over training steps.
(b) Student--teacher entropy gap over all tokens.
(c) Student--teacher entropy gap on selected tokens.
BiToK-SD maintains higher student entropy and larger student--teacher entropy gaps than the baselines, suggesting stronger policy exploration.
}
\label{fig:exploration_reflection}
\end{figure}

\begin{figure}[t]
\begin{center}
\includegraphics[width=\linewidth]{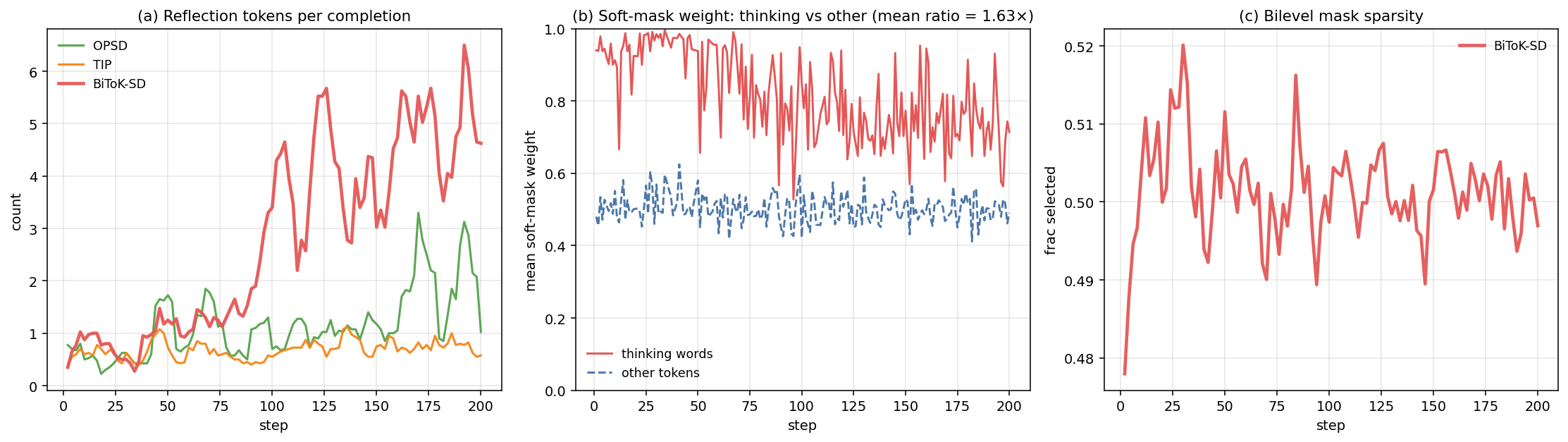}
\end{center}
\caption{Analysis of the bilevel token selection mechanism.
(a) Average number of epistemic markers per completion.
(b) Mean soft-mask weight on epistemic-marker positions
(e.g.\ \emph{wait}, \emph{hmm}, \emph{maybe}) versus all other token positions;
the average ratio is $1.63\times$.
(c) Fraction of tokens selected by the bilevel mask.
BiToK-SD produces more epistemic markers, and the mask assigns larger weights
to epistemic-marker positions while respecting the $k=50\%$ sparsity constraint.}
\label{fig:bilevel_selection_mechanism}
\end{figure}

To better understand where the improvement comes from, we analyze the training dynamics and the learned token-selection behavior in Figures~\ref{fig:exploration_reflection} and~\ref{fig:bilevel_selection_mechanism}.
Figure~\ref{fig:exploration_reflection} examines the exploration behavior of different methods during training.
In Figure~\ref{fig:exploration_reflection}(a), BiToK-SD maintains the highest student entropy for most training steps, indicating that the student policy preserves a broader output distribution rather than becoming overly concentrated.
Figure~\ref{fig:exploration_reflection}(b) further shows that BiToK-SD exhibits a larger student--teacher entropy gap over all tokens, suggesting stronger policy exploration relative to the fixed teacher.
The same trend is also observed for the selected tokens in Figure~\ref{fig:exploration_reflection}(c), where BiToK-SD consistently maintains a larger entropy gap than the TIP baseline.
Part of this contrast may also stem from TIP's reverse KL objective, whose
mode-seeking behavior drives the student to become sharper than its teacher.
The comparison with OPSD, which optimizes the same objective as BiToK-SD, indicates
that the bilevel-selected tokens are not merely used to enforce closer imitation of
the teacher; instead, they help preserve exploratory behavior during distillation.

Figure~\ref{fig:bilevel_selection_mechanism} provides a closer analysis of the learned bilevel token selection mechanism. Inspired by \citet{kim2026understanding}, we use a set of ten epistemic markers---\textit{wait}, \textit{hmm}, \textit{perhaps}, \textit{maybe}, \textit{actually}, \textit{alternatively}, \textit{seems}, \textit{might}, \textit{likely}, and \textit{check}---as lightweight lexical indicators of self-reflective behavior in generated completions. As shown in Figure~\ref{fig:bilevel_selection_mechanism}(a), BiToK-SD generates substantially more epistemic markers per completion than OPSD and TIP. The gap becomes visible around step~60 and continues to widen as training proceeds, which is consistent with the stronger exploration behavior observed in Figure~\ref{fig:exploration_reflection}.
To further examine how the selector allocates its weights, Figure~\ref{fig:bilevel_selection_mechanism}(b) compares the mean soft-mask weight assigned to positions containing epistemic markers with that assigned to all other valid token positions. Across training, reflection-associated positions receive consistently larger weights, with an average ratio of $1.63\times$. This contrasts with OPSD, which assigns uniform weights to all valid positions,
and TIP, which applies a binary Top-$K$ mask. In these baselines, the relative
weighting between reflection tokens and other selected tokens is uniform by
construction, whereas BiToK-SD assigns continuous weights that increase with the
utility score relative to an adaptive threshold. This indicates that the bilevel
selector not only determines \emph{which} tokens should contribute to the
distillation objective, but also controls \emph{how strongly} high-utility
positions, such as reflection-bearing ones, are emphasized.

Although BiToK-SD produces more reflection-style tokens, this does not translate into greater verbosity or degeneration. On AIME24 with 64 samples per problem, BiToK-SD generates 3.7\%, 1.6\%, and 4.8\%
fewer tokens than OPSD on Qwen3-1.7B, 4B, and 8B, respectively, while also showing
lower repeated 4-gram ratios and truncation rates at all three scales.
This suggests that the increased reflective behavior reflects a redistribution of the reasoning process rather than simply longer or more repetitive generation.
Finally, Figure~\ref{fig:bilevel_selection_mechanism}(c) shows that the overall mask density remains tightly concentrated around the target sparsity level $k=50\%$. Thus, the learned selector satisfies the global sparsity constraint without degenerating into either an overly sparse or overly dense mask. At the same time, the specific positions emphasized by the mask are determined adaptively from the training dynamics rather than imposed by a static token-selection heuristic.

Overall, the quantitative results and training analyses provide consistent evidence that BiToK-SD improves on-policy self-distillation through adaptive bilevel token selection. Beyond achieving stronger final-checkpoint performance, BiToK-SD also changes
the underlying distillation dynamics: the learned selector maintains the prescribed sparsity budget while preferentially emphasizing reflection-associated positions and preserving a broader student policy distribution. These results suggest that the benefit of BiToK-SD does not come from simply increasing the amount of distillation or uniformly enforcing teacher imitation, but from selectively applying token-level supervision to positions that are more useful for reasoning-oriented self-distillation. This adaptive balance between token selection and policy exploration yields a more effective mechanism for improving reasoning performance.

\subsection{Ablation Studies and Analysis}

We conduct ablation and sensitivity analyses to examine the design choices of
BiToK-SD. For computational efficiency, all ablations use Qwen3-1.7B and are
evaluated on AIME24 at the final checkpoint (200 steps) with 64 samples per
problem. Unless otherwise specified, all variants follow the same training setup. We additionally report Pass@64, defined as the
fraction of problems for which at least one of the 64 sampled solutions is correct.

\paragraph{Effect of the Top-$K$ ratio.}
We first vary $k \in {20\%,30\%,40\%,50\%,60\%,70\%}$ while keeping all
other settings fixed. As shown in Table~\ref{tab:ablation}, performance follows
a broadly inverted-U trend and peaks at the default $k=50\%$ (55.99 Avg@64 and
90.00 Pass@64). Nevertheless, performance remains competitive across the tested
range, indicating that BiToK-SD is reasonably robust to the choice of $k$.
The best performance at an intermediate selection ratio suggests a balance
between focusing distillation on informative tokens and retaining sufficient
coverage of the reasoning trajectory. Selecting too few tokens may discard
useful supervision, whereas emphasizing too many tokens may weaken the benefit
of token-level prioritization.

\begin{table}[t]
\centering
\small
\caption{Ablations on the Top-$K$ ratio and token-selection score on AIME24.}
\label{tab:ablation}
\setlength{\tabcolsep}{15pt}
\renewcommand{\arraystretch}{0.95}
\begin{tabular}{lcc@{\hspace{18pt}}lcc}
\toprule
\textbf{Top-$K$} & \textbf{Avg@64} & \textbf{Pass@64}
&
\textbf{Score} & \textbf{Avg@64} & \textbf{Pass@64} \\
\midrule
20\% & 53.59 & 83.33 & Uniform & 55.10 & 86.67 \\
30\% & 53.70 & 86.67 & Student Ent. & 54.79 & 83.33 \\
40\% & 54.22 & 80.00 & Teacher Ent. & 54.69 & 83.33 \\
\textbf{50\%} & \textbf{55.99} & \textbf{90.00} & KL & 55.78 & 86.67 \\
60\% & 54.95 & 76.67 & RKL & 53.85 & 83.33 \\
70\% & 54.53 & 86.67 & \textbf{Soft-OR} & \textbf{55.99} & \textbf{90.00} \\
\bottomrule
\end{tabular}
\end{table}

\paragraph{Effect of the token-selection score.}
We next examine whether BiToK-SD depends on a specific utility score.
We replace the default Soft-OR score with student entropy, teacher entropy,
forward KL divergence $D_{\mathrm{KL}}(P_T\|P_S)$, and reverse KL divergence
$D_{\mathrm{KL}}(P_S\|P_T)$ while fixing $k=50\%$. As a reference, we also
report uniform weighting over all valid tokens (Uniform), which recovers OPSD.
As shown in Table~\ref{tab:ablation}, the default Soft-OR score achieves the
highest Avg@64 and Pass@64 overall, while forward KL is the strongest
single-signal alternative. Both Soft-OR and forward KL improve Avg@64 over
uniform weighting, whereas the entropy-based scores and reverse KL do not.
The variation across utility scores remains within about two points in Avg@64,
suggesting that the proposed weighting framework remains reasonably robust to
the choice of utility score, although the score choice still affects the gain
over uniform weighting.

\paragraph{Effect of the mask temperature.}
We further vary the mask temperature
$\tau_m \in \{0.05,0.1,0.2\}$, obtaining Avg@64 scores of
55.00, 55.99, and 55.00 and Pass@64 scores of
80.00, 90.00, and 80.00, respectively.
The small variation indicates that BiToK-SD is not highly sensitive to the
precise choice of mask temperature.

\paragraph{Dynamics of the adaptive threshold.}
We further examine how the relaxed threshold $\rho$ evolves during training.
On Qwen3-1.7B, $\rho$ starts from 0.061, rises during the first third of training,
and returns to 0.063 by step 200 (range 0.044--0.105, mean $0.073 \pm 0.012$).
On the larger models, $\rho$ instead drifts upward throughout training: on
Qwen3-4B it increases from 0.030 to 0.102 (range 0.029--0.151, mean
$0.077 \pm 0.028$), and on Qwen3-8B from 0.069 to 0.159 (range 0.051--0.217, mean
$0.126 \pm 0.040$), with both peaks occurring late in training. These dynamics show
that the relaxed threshold adapts throughout training rather than remaining
effectively fixed.

\subsection{Generalization to Off-Policy KD}
\label{sec:off-policy}

To assess the generality of our bilevel token-selection framework, we further evaluate it under the off-policy distillation setting. Following the protocol of \citet{tavor2026rethinking}, we distill Qwen3-8B into Qwen3-1.7B on an 80M-token subset of FineWeb~\citep{penedo2024fineweb}, with a maximum sequence length of 512. Training runs for one epoch with an effective batch size of 64, using a micro-batch size of 2 and gradient accumulation of 8 on each of 4 GPUs, yielding roughly 2.4K optimizer updates. We use a learning rate of $1\times10^{-5}$, a KD temperature of $\tau=1.0$, and disable the cross-entropy term ($\alpha_{\text{ce}}=0$) so that all gradients originate from the distillation objective. To eliminate teacher forward passes during training, we precompute and cache, for every position, the teacher's top-64 vocabulary probabilities together with its entropy, which is required for the truncated distillation objective.

We compare against Full-KD, which distills all token positions; SE-KD, which selects token positions based on student entropy; and SE-KD\textsubscript{3X}, which extends selective distillation beyond the position axis by additionally applying selection along the class and sample axes, following \citet{tavor2026rethinking}. A practical constraint of this off-policy setup is that storing the teacher's full output distribution for every token position is prohibitively expensive. Therefore, only a truncated top-$U$ slice, with $U=64$, is retained on disk. As a result, statistics that require access to the complete teacher distribution---most notably the forward and reverse KL terms used by the Soft-OR importance score in our main experiments---cannot be evaluated faithfully. We therefore replace Soft-OR with the student-entropy score, which is computed solely from the student's predictions and is independent of the granularity of the cached teacher distribution. For the bilevel selector, we use the same relaxation hyperparameters as in the main experiments ($\tau_1=0.1$, $\tau_2=0.01$, $\tau_m=0.1$), and fix the position-selection ratio at $k=20\%$.

We evaluate on five downstream tasks---ARC-Easy, GSM8K, HellaSwag, PIQA,
and LAMBADA---and report the average accuracy (Avg.\ Acc.) over these
five tasks, the strict instruction-following score on the IFEval
benchmark, and the perplexity (PPL) on LAMBADA. As shown in
Table~\ref{tab:off-policy-results}, BiToK-SD outperforms Full-KD, SE-KD,
and SE-KD\textsubscript{3X} across all three summary metrics. These
results suggest that the proposed bilevel selection mechanism can also
be applied to off-policy KD, even when only truncated teacher
distributions are available.

\begin{table}[t]
\centering
\small
\caption{Off-policy knowledge distillation from Qwen3-8B (teacher) to
Qwen3-1.7B (student) using the FineWeb corpus as distillation data.
The first two rows show the undistilled student and teacher baselines.}
\setlength{\tabcolsep}{15pt}
\label{tab:off-policy-results}
\begin{center}
\begin{tabular}{lccc}
\toprule
Method
& Avg.\ Acc.\ $\uparrow$
& IFEval $\uparrow$
& PPL $\downarrow$ \\
\midrule
Qwen3-1.7B              & 61.90 & 19.40 & 12.20 \\
Qwen3-8B                & 73.80 & 29.00 & 4.60  \\
\midrule
Full-KD                 & 64.17 & 15.53 & 7.30  \\
SE-KD                   & 64.19 & 17.19 & 7.43  \\
SE-KD\textsubscript{3X} & 64.16 & 16.45 & 7.53  \\
\textbf{BiToK-SD}       & \textbf{64.77} & \textbf{18.67} & \textbf{6.94} \\
\bottomrule
\end{tabular}
\end{center}
\end{table}

\section{Conclusion}
We introduced BiToK-SD, a bilevel Top-$K$ token-selection framework for on-policy self-distillation in large language models.
Instead of distilling all generated tokens uniformly or relying on fixed heuristic selection rules, BiToK-SD learns adaptive token-level distillation weights through a differentiable threshold-based optimization formulation.
Experiments on mathematical reasoning benchmarks show that BiToK-SD
consistently improves over strong baselines while requiring only lightweight additional computation.
Further analyses indicate that the learned selector helps preserve policy exploration and is associated with more reflection-style reasoning behavior.
Additional ablation and off-policy KD experiments further demonstrate the robustness and generality of the proposed selection mechanism.
Overall, our results suggest that learning where and how strongly to apply distillation is a promising direction for improving self-distillation in reasoning-oriented language models.




\subsection*{AI use statement}


In this work, we used generative AI tools to assist with minor modifications to
our training and evaluation code. We have not used generative AI tools to
propose or refine hypotheses, design the methodology or experiments, interpret
results, or assist with translation. The remaining tasks requiring disclosure
are not applicable to this work. Additionally, we used generative AI tools to
improve the readability of the manuscript. We have reviewed all AI-assisted
work: all AI-modified code was reviewed and tested by the authors, and all
AI-edited text was checked against the original results for factual accuracy.
We take responsibility for the final content of this work, including text,
claims or artifacts produced with the aid of generative AI.

\subsection*{Reproducibility statement}


The complete specification of our method is given in Section~\ref{sec:method},
including the token-level utility score, the relaxed threshold objective, and
the soft mask and weighted distillation loss. Hyperparameters for the main
on-policy experiments---including the optimizer, learning rate, effective batch
size, LoRA configuration, decoding temperature, relaxation constants, and inner
solver settings---are listed in Section~4.1, with all baselines following the
same experimental configuration unless otherwise specified. The training data
is the mathematical reasoning subset of the publicly available OpenThoughts
corpus. The off-policy experiments in Section~\ref{sec:off-policy} use an
80M-token subset of the publicly available FineWeb corpus, with the training
protocol and teacher-distribution caching procedure described there. All models
in the main on-policy experiments are trained for a fixed 200 steps and
evaluated at the final checkpoint, whereas the off-policy experiments follow
the separate one-epoch protocol described in Section~\ref{sec:off-policy}.
Anonymized source code for reproducing our method and experiments is included
in the supplementary material.

\bibliography{iclr2027_conference}
\bibliographystyle{iclr2027_conference}


\end{document}